%% file: neurips_2023.tex
\documentclass{article}

\PassOptionsToPackage{numbers, compress}{natbib}
\usepackage[preprint]{neurips_2023}
\usepackage[utf8]{inputenc} % allow utf-8 input
\usepackage[T1]{fontenc}    % use 8-bit T1 fonts
\usepackage{hyperref}       % hyperlinks
\usepackage{url}            % simple URL typesetting
\usepackage{booktabs}       % professional-quality tables
\usepackage{amsfonts}       % blackboard math symbols
\usepackage{nicefrac}       % compact symbols for 1/2, etc.
\usepackage{microtype}      % microtypography
\usepackage{xcolor}         % colors
\usepackage{wrapfig}
\usepackage{bbm}
\usepackage[utf8]{inputenc} % allow utf-8 input
\usepackage[T1]{fontenc}    % use 8-bit T1 fonts
\usepackage{hyperref}       % hyperlinks
\usepackage{url}            % simple URL typesetting
\usepackage{booktabs}       % professional-quality tables
\usepackage{amsfonts}       % blackboard math symbols
\usepackage{nicefrac}       % compact symbols for 1/2, etc.
\usepackage{microtype}      % microtypography
\usepackage{algorithm}
\usepackage{algorithmic}
\usepackage{mathtools}
\usepackage{amsthm}
\usepackage{amssymb}% http://ctan.org/pkg/amssymb
\usepackage{amsmath}
\usepackage{pifont}

\usepackage[capitalize,noabbrev]{cleveref}
\usepackage{thmtools, thm-restate}
\theoremstyle{plain}
\newtheorem{theorem}{Theorem}[section]

\theoremstyle{definition}

\theoremstyle{remark}
\newtheorem{remark}[theorem]{Remark}
\usepackage[textsize=tiny]{todonotes}
\usepackage{microtype}
\usepackage{graphicx}
\usepackage{subfigure}
\usepackage{natbib}
\usepackage{booktabs} % for professional tables
\usepackage{enumitem}
\usepackage{titletoc}

\title{Reinforcement Learning in the Era of LLMs: \\
What is Essential? What is needed? \\
An RL Perspective on RLHF, Prompting, and Beyond.}

\author{%
  Hao Sun\\
  Department of Applied Mathematics and Theoretical Physics\\
  University of Cambridge\\
  \texttt{hs789@cam.ac.uk}\\
}

\begin{document}

\maketitle

\begin{abstract}
 Recent advancements in Large Language Models (LLMs) have garnered wide attention and led to successful products such as ChatGPT and GPT-4. Their proficiency in adhering to instructions and delivering harmless, helpful, and honest (3H) responses can largely be attributed to the technique of Reinforcement Learning from Human Feedback (RLHF). In this paper, we aim to link the research in conventional RL to RL techniques used in LLM research. Demystify this technique by discussing why, when, and how RL excels. Furthermore, we explore potential future avenues that could either benefit from or contribute to RLHF research.
 
 \textbf{\textit{Highlighted Takeaways}}:
 \begin{enumerate}
     \item RLHF is Online Inverse RL with Offline Demonstration Data. 
     \item RLHF $>$ SFT because Imitation Learning (and Inverse RL) $>$ Behavior Cloning (BC) by alleviating the problem of compounding error.
     \item The RM step in RLHF generates a proxy of the expensive human feedback, such an insight can be generalized to other LLM tasks such as \textit{prompting} evaluation and optimization where feedback is also expensive. 
     \item The policy learning in RLHF is more challenging than conventional problems studied in IRL due to their high action dimensionality and feedback sparsity.
     \item The main superiority of PPO over off-policy value-based methods is its stability gained from (almost) on-policy data and conservative policy updates.
 \end{enumerate}
\end{abstract}

\input{tex/introduction_to_rl}

\bibliography{ref}
\bibliographystyle{unsrtnat}

\end{document}

%% file: tex/introduction_to_rl.tex
\section{A Crash Introduction to RL: Online RL, Offline RL, and Inverse RL}

In this section, we will briefly introduce some basic concepts needed in our discussion later. We begin by highlighting the important intuitions behind the technique of Reinforcement Learning (RL), followed by a more technical formalism. Our goal is to ensure everyone, regardless of their background, can grasp the intricacies of RL and its impact on Large Language Models.

\subsection{Essential Concepts}
In RL, an agent learns through interacting with an environment and receiving feedback in the form of rewards. The fundamental objective of RL is to find a policy, which is a mapping from states to actions, that maximizes the expected cumulative reward over time.

Here are several useful concepts:
\begin{itemize}
    \item (\textbf{Environment} \includegraphics[height=1em]{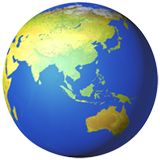}=\includegraphics[height=1em]{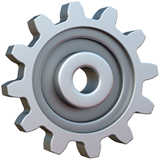}+\includegraphics[height=1em]{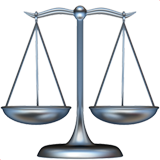}.) When we are talking about an \textit{Environment} \includegraphics[height=1em]{figs/earth_asia.png} we are talking about the \textit{dynamics model }\includegraphics[height=1em]{figs/gear.png} and the \textit{reward function} \includegraphics[height=1em]{figs/scales.png}. 
    \item (\textbf{Agent} \includegraphics[height=1em]{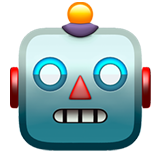}) An \textit{Agent} is the subject of a \textit{policy} that interacts with the environment. In a sequential decision-making problem, there can be multiple decision steps, and a smart policy will make its decision at every step by considering every piece of information it can collect till then. e.g., using recurrent networks to record histories in \cite{sun2021safe,ni2021recurrent}
    \item (\textbf{Difficulties} \includegraphics[height=1em]{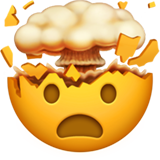}) Why is it hard to learn? 1. the learning objective is non-differentiable, it engages the unknown environment. 2. the policy needs to trade off between exploring \textit{random} novel behaviors that potentially can be better than the current, yet as those are random behaviors, they are usually worse than the current --- you may imagine how hard it would be for the LLM generation tasks when there are ~10k tokens (as action) to choose from ...
    \item (\textbf{Learning} \includegraphics[height=1em]{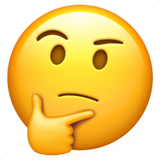}) The key insight behind the \textit{learning} step in RL is to increase the probability of executing the \textit{good} actions (which leads to a high cumulative future reward) and decrease the probability of executing \textit{bad} actions (which have a low cumulative future reward). An easy-to-follow approach can be performing supervised learning on a collected set of good actions. e.g., Using Supervised Learning to mimic successful trajectories as an alternative approach to RL~\cite{sun2019policy,sun2020zeroth}.
\end{itemize}

\subsection{Technical Formumation}
RL can be formally represented using the Markov Decision Processes (MDPs), where decisions are made in discrete time steps, and each decision affects the state of the environment in the subsequent step.
\subsubsection{Markov Decision Processes}
Formally, we denote the MDP as $\mathcal{M} = \{\mathcal{S},\mathcal{A},\mathcal{T},\mathcal{R},\rho_0,\gamma\}$, where $\mathcal{S}\subset \mathbb{R}^{d}$ denotes the $d$-dim state space, $\mathcal{A}$ is the action space. Broadly, the environment includes $\mathcal{T}$ and $\mathcal{R}$, the former denotes the transition dynamics $\mathcal{T}: \mathcal{S}\times \mathcal{A} \mapsto \Delta(\mathcal{S})$ that controls transitions between states, and the reward function $\mathcal{R}:\mathcal{S}\times\mathcal{A}\mapsto \mathbb{R}$ provides feedback. In the most common settings, we assume the feedback is a scalar, yet in risk-sensitive or cost-sensitive settings, the reward function can be a vector, where constrained optimization techniques can be applied~\cite{sun2020novel,sun2022constrained}. $\rho_0 = p(s_0)\in\Delta(\mathcal{S})$ denotes the initial state distribution. $\gamma$ is the discount factor that trades off between short-term and long-term returns.
\subsubsection{Online RL}
\begin{figure}[h!]
    \centering
    \includegraphics[width=0.5\linewidth]{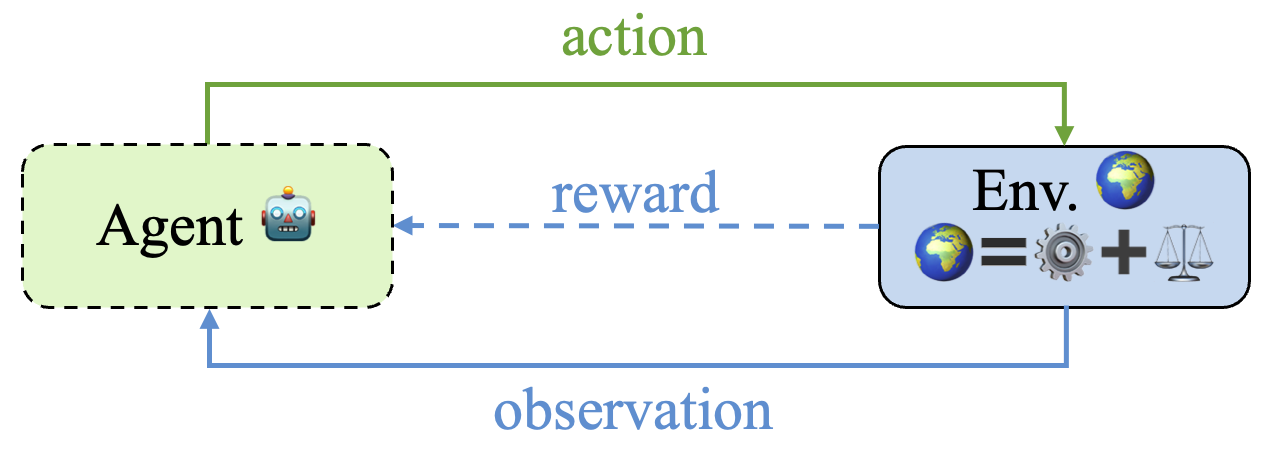}
    \caption{\small A pictorial illustration of RL: an agent interacts with the environment and learns from trial and error.}
    \label{fig:1_onlinerl}
\end{figure}
In the \textit{Online RL} setting, an agent with policy $\pi\in\Pi:\mathcal{S}\mapsto \Delta (\mathcal{A})$ learns through trial and error. It actively interacts with the environments --- including both transition dynamics $\mathcal{T}$ and the reward function $\mathcal{R}$. 

At each time step $t$, an agent observes a state $s_t$ from the environment and selects an action $a_t \sim \pi$. Upon taking the action, the agent receives a reward $r_t$ and transit to a new state $s_{t+1}$. The agent's objective is to maximize its expected return.
\begin{equation}
    \pi^* = \arg\max_{\pi\in\Pi}\mathbb{E}_{a_t\sim\pi, s_{t+1}\sim \mathcal{T},s_0\sim \rho_0}\sum_{t=0}^T \gamma^t \mathcal{R}(s_t, a_t),
\end{equation}
We can alternatively denote the trajectory generated by a policy $\pi$ to be $\tau = \{s_0, a_0\sim\pi(a_0|s_0),s_1\sim\mathcal{T}(s_1|s_0, a_0), a_1\sim \pi(a_1|s_1),... \}$ and denote the trajectory distribution of $\pi$ as
\begin{equation}
p_\pi(\tau) =\rho_0 \Pi_{t=0}^{T} \pi(a_t|s_t)\mathcal{T}(s_{t+1}|s_t,a_t), 
\end{equation}
where $T$ denotes the length of decision sequences. The learning objective can be expressed as 
\begin{equation}
\pi^* = \arg\max_\pi \mathbb{E}_{\tau\sim p_\pi (\tau)}\left[\sum_{t=0}^{T} \gamma^t \mathcal{R}(s_t, a_t) \right].
\end{equation}
\subsubsection{Offline RL}
In the \textit{Offline RL} setting, interactions with the environment are strictly forbidden. The learning problem is no longer online learning but learning from a static dataset of decision logs $\mathcal{D}_{\mathrm{Off-RL}} = \{(s^i_t,a^i_t,s^i_{t+1},r^i_t)\}$, that is generated by some unknown behavior policy $\pi_\beta$.

The most obvious difficulty in the offline RL setting is such a setting prohibits exploration --- hence it hinders the improvement of policy learning to be improved over the demonstration data (though sometimes the learned policy can be better than the demonstration). 

Another fundamental challenge is the \textit{distributional shift}: although offline RL learns from a static dataset, its evaluation is actually based on rolling out a policy in an environment --- this is different from the ordinary supervised learning settings where the training set and test set are sampled from the same distribution. In offline RL training, the state distribution is sampled from rolling out the behavior policy $\pi_\beta$, whereas in its evaluation, the state distribution is sampled from rolling out the learned policy $\pi$.

To be more specific, assuming the decision dataset is collected from an optimal behavior policy $\pi_\beta^*$, such that every decision $a^i_t$ is optimal. We denote the state-action pairs in the dataset as $(s_t, a^*_t)$, then the expected number of mistakes made by the learned policy $\pi$ based on such an expert decision dataset can be denoted as
\begin{equation}
\ell(\pi) = \mathbb{E}_{p_\pi(\tau)} \left[ \sum_{t=0}^T \mathbbm{1}(\pi(s_t)\ne a^*_t) \right]    
\end{equation}
Then we have the following theorems:
\begin{theorem}[Behavior Clone Error Bound. \citet{ross2011reduction}]
\label{theorem:1}
     If $\pi$ is trained via empirical risk minimization on $s_t\sim p_{\pi_\beta}(\tau)$ and optimal labels $a_t^*$, and attains generalization error $\epsilon$ on $s_t \sim p_{\pi_\beta}(\tau)$, then $\ell(\pi)\le C+T^2 \epsilon$ is the best possible bound on the expected error of the learned policy.
\end{theorem}
\begin{remark}[Compounding Error.]
    An intuitive interpretation of this quadratic relationship between the error bound and the generalization error is that those errors aggregate along the trajectory. i.e., whenever the learned policy makes a mistake, it tends to make more mistakes from then on as that action is not optimal and will lead to other out-of-distribution states, which will lead to further mistakes.
\end{remark}
\begin{remark}[Behavior Clone]
    We can always set up a supervised learning objective in offline RL to minimize the difference between decision demonstration pairs. i.e.,
\begin{equation}
    \pi = \arg\min_\pi \mathbb{E}_{(s^i_t,a^i_t)\sim\mathcal{D}} ||a^i_t -\pi(s^i_t)||^2
\end{equation}
\end{remark}
\begin{figure}[h!]
    \centering
    \includegraphics[width=0.8\linewidth]{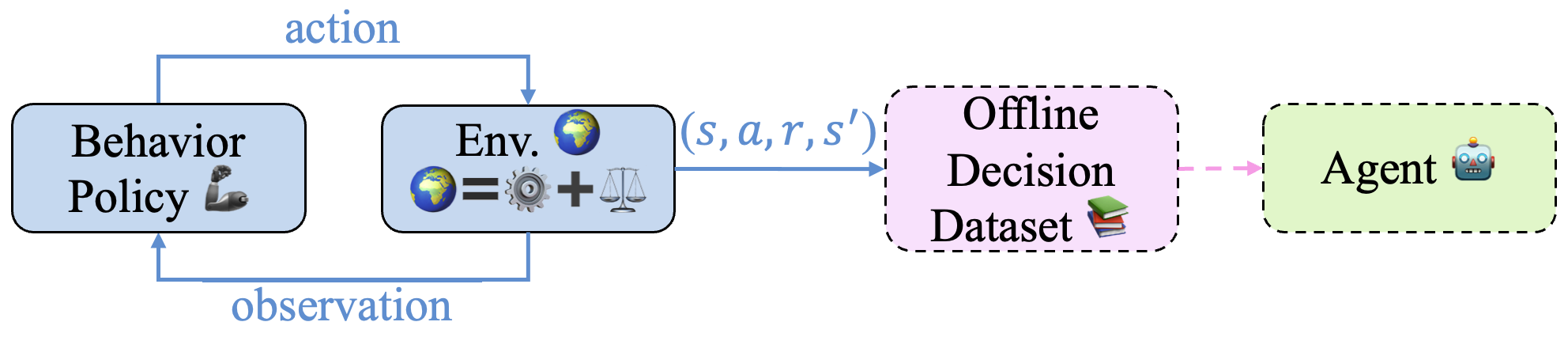}
    \caption{\small In Offline RL, a behavior policy interacts with the environment and generates a decision dataset. Then such a decision dataset is used to learn a policy without access to the environment (offline).}
    \label{fig:2_offlinerl}
\end{figure}
\subsubsection{Imitation Learning}

In order to alleviate the challenge of compounding error we discussed above, \textit{Imitation Learning} (IL) considers the setting where a dynamics model is available during learning. 

\paragraph{Another Motivation of IL: Reward Design is Hard} The setup of IL is especially common for problems where reward engineering is hard. This is because although the “reward hypothesis” tells us whenever we can define a reward function for a task, it can be solved by RL, it does not consider whether this task can be efficiently solved. For instance, in playing Go or StarCraft, it's easy to define a reward function that returns $+1$ when winning and $0$ when losing. However, it will not be hard to imagine that such a reward function is extremely sparse to provide helpful information during learning. In another example of teaching robots to finish complex tasks, imitation can also circumvent the difficulty of describing a motion sequence with a reward function~\cite{peng2018deepmimic}.

\paragraph{A Method for Reward Engineering} In a previous paper~\cite{sun2022exploit}, we show and illustrate why using a $0$ for win and $-1$ for lose is better than using  $+1/0$. A simple reward shifting with a few lines of code added to the RL reward function can be used to improve exploration (for Online RL) or enhance conservative exploitation (for Offline RL).

To alleviate the challenge of reward engineering in RL tasks, IL is introduced to learn to use the dynamics model but without a pre-defined reward model. Consider those examples: (1) in learning humanoid robotics locomotion skills, it is hard to define an objective to let the robot “walk as a human” --- however, providing demonstration data to show how humans walk is much easier. (2) in autonomous driving, it is hard to define the objective of “driving safe and well” --- however, we should be able to provide human driving videos or control sequences as demonstrations of good and safe driving behaviors.

The objective of IL is to learn from a (decision) demonstration dataset, with access to a dynamics model --- such that the current policy can be rolled out in the real environment. Intuitively, with such a dynamics model, the optimization objective will no longer be $s_t\sim p_{\pi_\beta}(\tau)$ but could be $s_t\sim p_{\pi}(\tau)$ --- the distributional shift problem can be alleviated.

\begin{figure}[h!]
    \centering
    \includegraphics[width=0.6\linewidth]{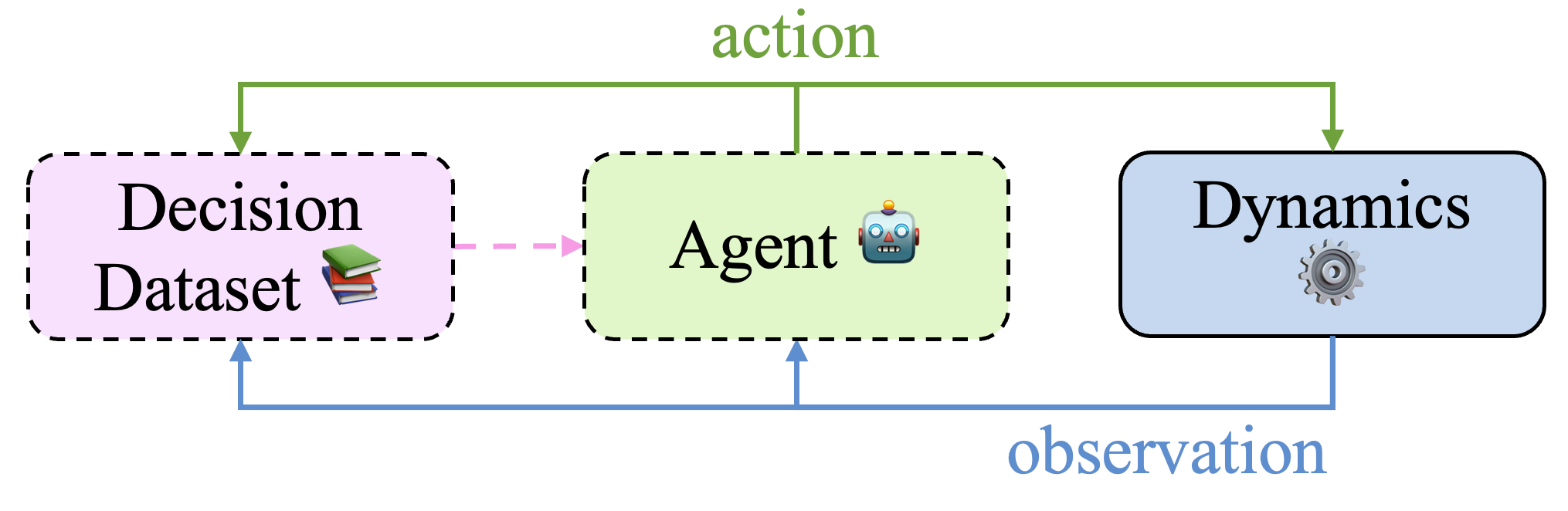}
    \caption{\small In Imitation Learning (IL), the agent learns from feedback from the decision dataset, but the observations are from a real dynamics model.}
    \label{fig:3_il}
\end{figure}

There are many practical methods for implementing such a learning process, and the most famous work in the Deep-RL era is the GAIL~\citep{ho2016generative}, which conducts IL through adversarial learning: the policy is a \textit{generator} of behaviors, while a \textit{discriminator} then tries to identify whether a trajectory is generated by the behavior policy $\pi_\beta$ or by the generator (the policy learned).

For the theory results, we have the following theorem:

\begin{theorem}[DAgger Error Bound, \citet{ross2011reduction}]
\label{theorem:2}
    If $\pi$ is trained via empirical risk minimization on $s_t\sim p_{\pi}(\tau)$ and optimal labels $a_t^*$, and attains generalization error $\epsilon$ on $s_t\sim p_{\pi}(\tau)$, then $\ell(\pi)\le C+T \epsilon$ is the best possible bound on the expected error of the learned policy.
\end{theorem}
\begin{remark}
    This requires the additional assumption of being able to access the behavior (expert) policy $\pi_\beta$ actively to acquire the expert for those roll-out trajectories generated by $\pi$ . 
\end{remark}

\textbf{Takeaway:} Comparing Theorem \ref{theorem:1} and Theorem \ref{theorem:2}, we see that \textbf{having access to a \textit{dynamics model} \includegraphics[height=1em]{figs/gear.png} is essential in controlling the error bound.}

\subsubsection{Inverse Reinforcement Learning}

Inverse reinforcement learning (IRL) is actually just one of the many solutions to IL problems, with an emphasis on reward model learning. It first learns a reward model, and then uses such a reward model --- combined with the dynamics model --- to perform online RL.
\begin{figure}[h!]
    \centering
    \includegraphics[width=0.85\linewidth]{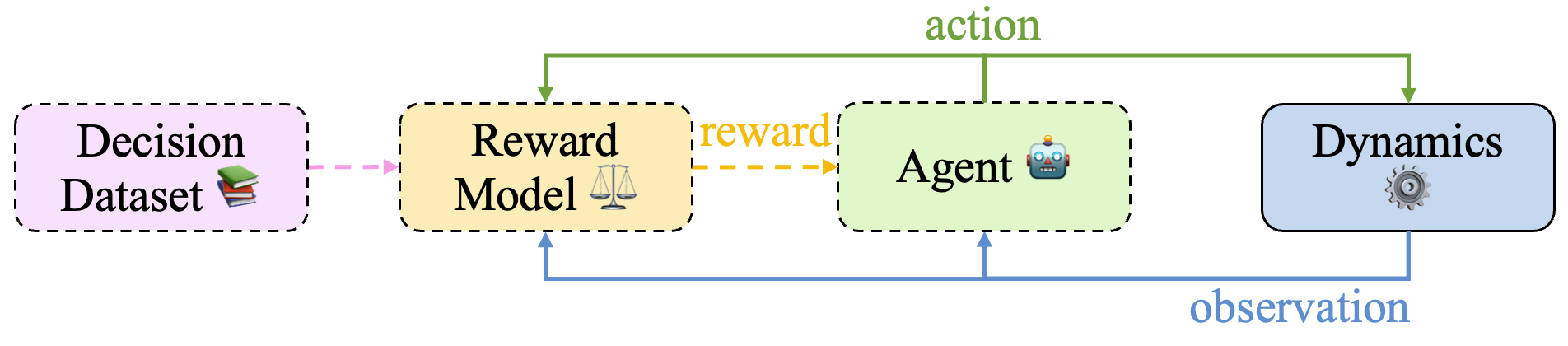}
    \caption{\small Inverse Reinforcement Learning (IRL) solves the IL tasks in two steps: (1). reward modeling that distills the knowledge of underlying learning objectives that the behavior policy seems to optimize from the offline decision demonstration dataset. (2). combining such a learned reward model and the accessible dynamics model, everything needed for an online RL algorithm is right there.}
    \label{fig:4_irl}
\end{figure}

\textbf{Offline IL and Offline IRL:} What if both the reward model and dynamics model are not available? This situation is clearly more challenging. The demonstration dataset in such settings will be in the form of $\mathcal{D}_{\mathrm{OIL}} = \{(s^i_t,a^i_t,s^i_{t+1})\}$. Besides the behavior cloning method, there are several alternative approaches like the energy-based method SBIL~\cite{jarrett2020strictly}, and the latent space decomposition method ABC~\cite{sun2023accountable}. ABC can be regarded as an accountable counterpart of BC, therefore, it works in all settings where BC can be applied.

\subsubsection{Learning from Demonstrations} Another related but different topic is Learning from Demonstrations (LfD)~\cite{schaal1996learning,hester2018deep,nair2018overcoming}, which leverages the demonstration dataset as a warm-start for RL. For instance, in the aforementioned tasks of Go or StarCraft, we can first use the demonstration dataset to perform behavior cloning (BC) and then use the learned BC policy as a warm start for RL. LfD also benefits the exploration of robotics control tasks where the reward can be extremely sparse, and defined as “whether the goal is achieved”. In a nutshell, LfD uses demonstrations to improve exploration in reward sparse tasks, and those demonstrations may not be optimal (e.g., non-expert players’ replay of StarCraft~\cite{vinyals2019grandmaster}), LfD then returns to RL and a sparse reward function to further refine the policy learned from demonstration dataset.

\subsubsection{Comparison Between Different Settings}
The table below summarizes the differences between RL, Offline-RL, IL, IRL, Offline-IRL, and LfD.
\begin{table}[h]
\fontsize{8}{10}\selectfont
\centering
\caption{\small Summarization of the differences between RL, Offline-RL, IL, IRL, Offline-IRL, and LfD.}
\begin{tabular}{l|c|c|c|c|c}
\toprule
\textbf{Problem} & \textbf{External} & \textbf{External} & \textbf{Learned} & \textbf{(Near)-Expert} & \textbf{Example} \\
\textbf{Settings} & \textbf{Dynamics} & \textbf{Reward} & \textbf{ Reward} & \textbf{Demonstration} & \textbf{Solvers} \\
\textbf{} & \textbf{Model} & \textbf{Model} & \textbf{Model} & \textbf{} & \\
\midrule
RL & \includegraphics[height=1em]{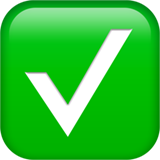} & \includegraphics[height=1em]{figs/white_check_mark.png} & \includegraphics[height=1em]{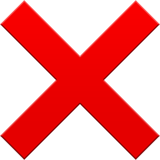} & \includegraphics[height=1em]{figs/x.png} & PPO~\cite{schulman2017proximal}, TD3~\cite{fujimoto2018addressing},SAC~\cite{haarnoja2018soft}\\
% \hline
Offline-RL & \includegraphics[height=1em]{figs/x.png} & \includegraphics[height=1em]{figs/x.png} & \includegraphics[height=1em]{figs/white_check_mark.png} or \includegraphics[height=1em]{figs/x.png} & \includegraphics[height=1em]{figs/white_check_mark.png} & BC, ABC~\cite{sun2023accountable}, CQL~\cite{kumar2020conservative}, WGCSL~\cite{yang2022rethinking} \\
% \hline
IL & \includegraphics[height=1em]{figs/white_check_mark.png} & \includegraphics[height=1em]{figs/x.png} & \includegraphics[height=1em]{figs/white_check_mark.png} or \includegraphics[height=1em]{figs/x.png} & \includegraphics[height=1em]{figs/white_check_mark.png} & BC, ABC~\cite{sun2023accountable}, GAIL~\cite{ho2016generative} \\
% \hline
IRL & \includegraphics[height=1em]{figs/white_check_mark.png} & \includegraphics[height=1em]{figs/x.png} & \includegraphics[height=1em]{figs/white_check_mark.png} & \includegraphics[height=1em]{figs/white_check_mark.png} & BC, ABC~\cite{sun2023accountable}, T-REX~\cite{brown2019extrapolating} \\
% \hline
Offline-IRL & \includegraphics[height=1em]{figs/x.png} & \includegraphics[height=1em]{figs/x.png} & \includegraphics[height=1em]{figs/white_check_mark.png} & \includegraphics[height=1em]{figs/white_check_mark.png} & BC, ABC~\cite{sun2023accountable}, SBIL~\cite{jarrett2020strictly} \\
% \hline
LfD & \includegraphics[height=1em]{figs/white_check_mark.png} & \includegraphics[height=1em]{figs/white_check_mark.png} & \includegraphics[height=1em]{figs/x.png} & \includegraphics[height=1em]{figs/white_check_mark.png} & DQNfD~\cite{hester2018deep}, DDPGfD~\cite{nair2018overcoming}, AlphaStar~\cite{vinyals2019grandmaster} \\
\bottomrule
\end{tabular}
\end{table}
\section{RLHF: Solving the Problem of Offline RL with Online Inverse RL}

\subsection{LLM Alignment from Human Feedback}

In the task of LLM alignment from human feedback, LLMs are fine-tuned to better follow user instructions. In the seminal paper of OpenAI~\cite{ouyang2022training}, such an alignment includes two general parts: supervised fine-tuning (SFT) and reinforcement learning from human feedback (RLHF). Figure~\ref{fig:5_rlhf} below illustrates those concrete steps. The first part of SFT is relatively easy to follow and implement, yet the secret and insight behind RLHF are more intricate.

\begin{figure}[h!]
    \centering
    \includegraphics[width=0.9\linewidth]{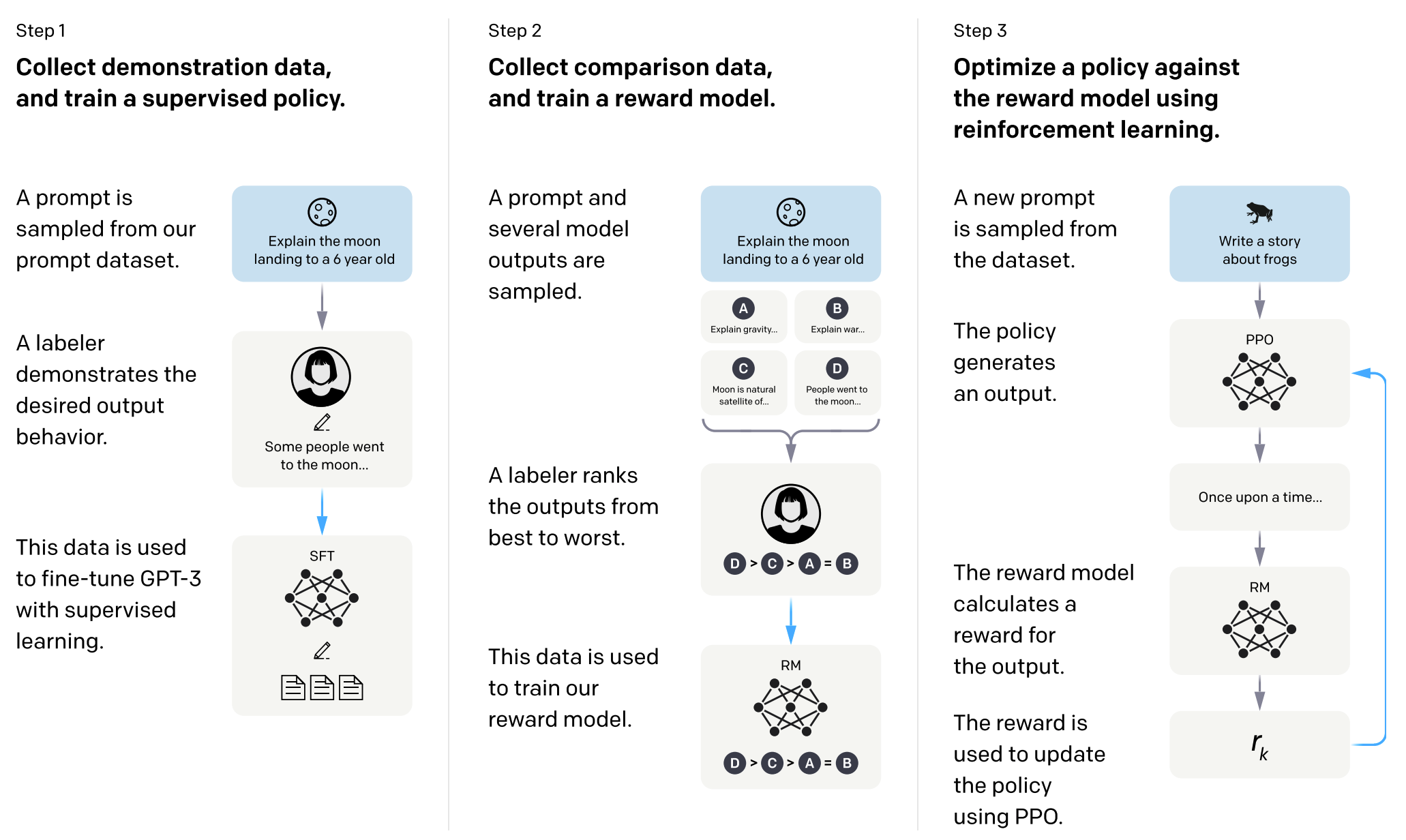}
    \caption{\small (From \citet{ouyang2022training}) There are 3 steps to align LLMs to human preference. Step 1: supervised fine-tuning of pre-trained LLM to follow instructions (generated by human demonstration data). Step 2: sample multiple responses for every query, and rank those responses according to human preference. Then a reward model can be learned to mimic the human preference. Step 3: Optimize the language model through RL to maximize the feedback from the reward model}
    \label{fig:5_rlhf}
\end{figure}

\subsection{Aligning with Human Preference: the Online Nature and Offline Practice}

\begin{figure}[h!]
    \centering
    \includegraphics[width=0.66\linewidth]{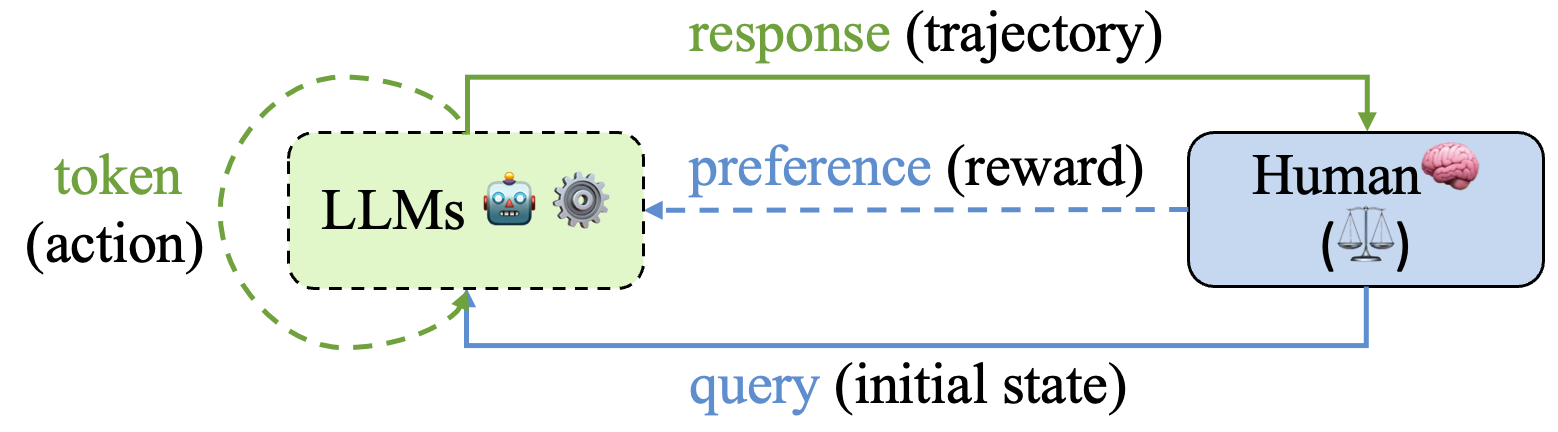}
    \caption{\small RLHF as an online RLproblem: Human preference is the underlying reward model, however, querying humans to provide feedback on every response is usually infeasible.}
    \label{fig:6_rlhf_online}
\end{figure}

Ideally, the RLHF phase can be conducted with human-in-the-loop, as shown in Figure~\ref{fig:6_rlhf_online}. In such an online setting, Human provides feedback to every response of LLMs, and LLMs learn from the external reward model of human preference. In fact, OpenAI now should be able to conduct such a process by collecting user’s feedback on ChatGPT’s response. But usually, such an online setting is infeasible due to its high cost of keeping humans in the loop.

\begin{figure}[h!]
    \centering
    \includegraphics[width=0.8\linewidth]{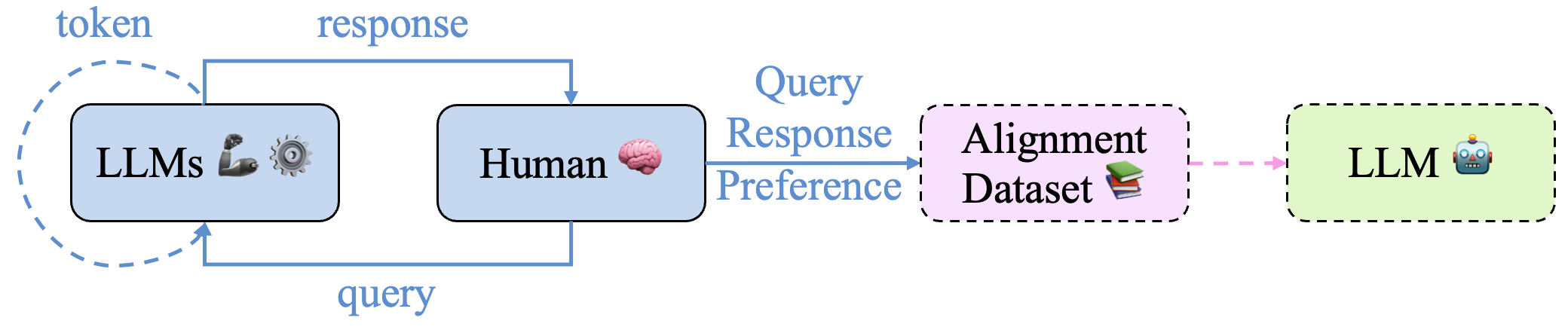}
    \caption{\small Because of the high cost of keeping humans in the loop, the practice of RLHF considers learning with an offline dataset generated by interactions between (the SFT) LLMs and Human annotators. The generated offline dataset is then used for LLM alignment.}
    \label{fig:7_rlhf_offline}
\end{figure}

Practically, RLHF addresses such a difficulty by generating an \textit{offline alignment dataset} that contains different queries (i.e., states $s$ in RL), responses (i.e., trajectory $\tau$ in RL), and preferences provided by human annotators (i.e., reward $r$ in RL). From such a perspective, the RLHF may seem to be a natural online RL problem but adjusted into \textit{an offline RL problem} due to cost considerations. Figure~\ref{fig:7_rlhf_offline} illustrates such a generation process.

Recall the problems of distributional shift and compounding error we discussed above in Offline RL, it seems RLHF must suffer from such problems. However, we show in the next section that RLHF can actually be solved as an \textbf{Online IL} problem, rather than an offline RL problem.

\subsection{RLHF: From Offline-RL to Online Imitation}

The essential observation we would highlight in RLHF is that \textbf{\textit{the dynamics model in response generation is known.}} Specifically, harking back to Figure 6, the actions are tokens generated by LLMs, and the responses (trajectories) are concatenations of those generated tokens. Given the auto-regression nature of LLMs’ generation process, given an initial query denoted as $s_0$, we can formally write the trajectory generation process of an LLM $\ell$ as follows:
\begin{itemize}
    \item $s_0 \sim p(s_0)$: (\textit{Interpretation:}) sample from query distribution | (\textit{RL Language:}) sample from initial state distribution
    \item $a_0 \sim \ell(s_0)$: (\textit{Interpretation:}) sample the next token with $\ell$ | (\textit{RL Language:}) sample action from the policy
    \item $s_1 = \mathcal{T}(s_0,a_0) = \mathrm{Concat}(s_0, a_0)$: (\textit{Interpretation:}) concatenate the generated token and the query as input for LLM for the next token generation | (\textit{RL Language:}) the transition dynamics gives the next state
    \item $a_1\sim \ell(s_1)$: ...
    \item ...
\end{itemize}

\begin{figure}[h!]
    \centering
    \includegraphics[width=0.4\linewidth]{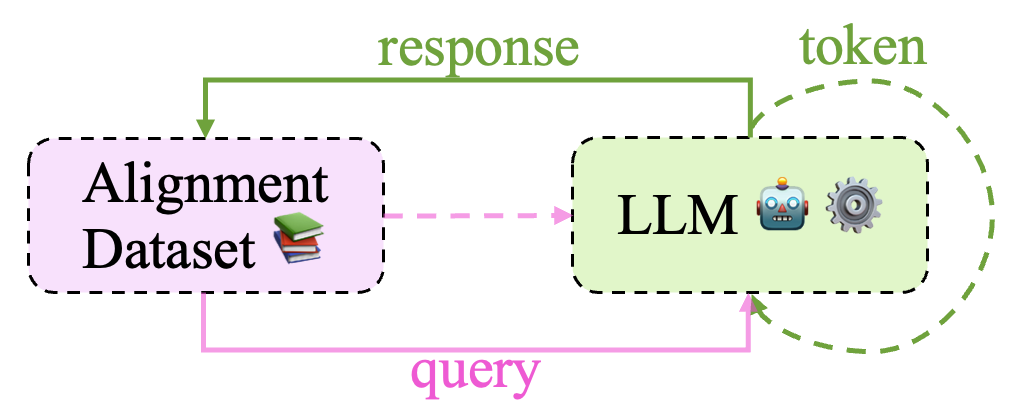}
    \caption{\small When aligning LLMs using an offline dataset, the dynamics model is a concatenation of the generated token and existing tokens, therefore, the offline RL problem can be solved by online IL.}
    \label{fig:8_rlhf_onlineil}
\end{figure}

Figure~\ref{fig:8_rlhf_onlineil} showcases why the LLM alignment task can be solved as an online IL in practice (c.f. Figure~\ref{fig:3_il}: pictorial illustration of IL)

\begin{figure}[h!]
    \centering
    \includegraphics[width=0.6\linewidth]{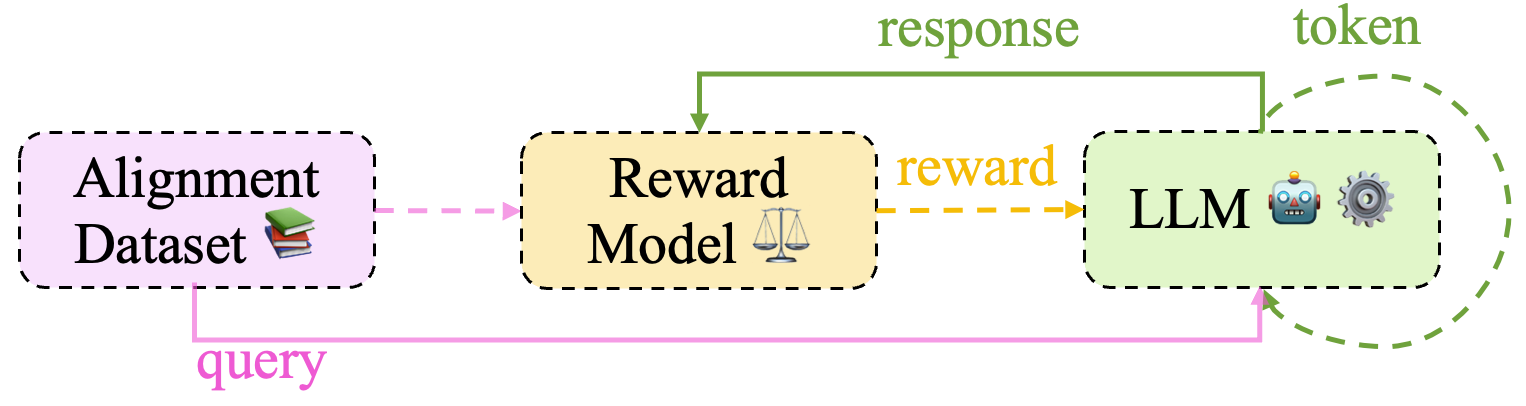}
    \caption{\small RLHF is online IRL. The reward modeling step learns a reward function from the alignment dataset, and given such a reward model and the known transition dynamics (concatenation), online RL algorithms like PPO can then be applied.}
    \label{fig:9_rlhf_onlineirl}
\end{figure}

Practically, RLHF chooses to use the Inverse RL approach for the IL problem --- with the first step explicitly learning a reward model, and the second step conducting RL using such a reward model. Figure~\ref{fig:9_rlhf_onlineirl} illustrates the learning procedure.

\textbf{Takeaway:} When aligning LLMs with an \textit{\textbf{offline human-preference alignment dataset}}, RLHF uses an \textit{\textbf{online IRL}} approach. This is because the transition dynamics model is known. Leveraging such a property, \textbf{the compounding error and distributional shift problems of offline RL can be alleviated.}

\subsection{Challenges and Open Questions from an RL Perspective}

\subsubsection{Why is RLHF better than SFT?}

Given the discussions above, the reason RLHF can be better than SFT --- from an RL perspective --- is that RLHF leverages the fact that the dynamics model is known, and uses IL to solve the alignment problem. On the other hand, SFT corresponds to the behavior clone approach, which suffers from the problem of compounding error. 
\textbf{Therefore, as long as IL > BC, we have RLHF > SFT.}

\subsubsection{Is PPO the Only Solution?}

Recently, several works have proposed alternatives to RLHF, including DPO~\cite{rafailov2023direct} that directly optimizes the LLM using human-preference data without reward modeling, RRHF~\cite{yuan2023rrhf} and RAFT~\cite{dong2023raft} propose ranking-based sampling methods as alternatives to PPO to address the computational instability issue and high GPU memory demand of PPO.

So clearly, there is still a lot of space for future improvement over PPO. We would like to mention the following pros and cons (mostly based on empirical observations) of PPO:
\begin{enumerate}
    \item PPO works well in large-scale discrete tasks~\cite{vinyals2019grandmaster}. The action space of LLM is far larger than normal RL problems.
    \item PPO has a faster wall-clock training time compared to off-policy methods like DQN~\cite{mnih2013playing}. PPO can be highly environment-parallelized. In fact, this is normally an implementation problem: in DQN a higher Update-to-Data (UTD) ratio~\cite{janner2019trust} is used: updates of networks in DQN are conducted every time step, but in PPO the updates of networks only happen at the end of an episode. e.g., after the entire response is generated.
    \item Aligning to human preference is a sparse reward problem. In the sense that only at the end of an episode will the agent receive a reward signal (provided by human feedback or the learned reward function). Such a setting is relevant to the multi-goal robotics control tasks~\cite{plappert2018multi} where the idea of Hindsight learning shines with the value-based methods~\cite{andrychowicz2017hindsight,sun2019policy} --- rather than policy-based methods like TRPO~\cite{schulman2015trust} and PPO. There are several attempts using the Hindsight relabeling trick for LLM fine-tuning~\cite{liu2023languages,zhang2023wisdom}.
    \item A fun fact is that policy-gradient and value-based methods are almost equivalent~\cite{schulman2017equivalence}. But in practice, the studies on LLM finetuning now mainly focus on the on-policy policy-based methods. The performance differences between policy-based methods and value-based methods can be mainly attributed to (1). on-policy/ off-policy data --- the staleness of the data they used for value and policy learning; and (2). whether using an aggressive and explicit or conservative and implicit policy learning --- while the policy-gradient methods like PPO and TRPO use a value function \textit{implicitly} for policy learning (i.e., use them as \textit{critics} to calculate policy gradient values that improve policy quality), the value-based methods like TD3 and SAC \textit{explicitly} turns the learned value function into policies (i.e., through either deterministic policy gradient DPG~\cite{silver2014deterministic} in TD3 or the Boltzmann policy~\cite{o2016combining} as in SAC/soft Q-learning~\cite{haarnoja2017reinforcement}.)
\end{enumerate}

\subsubsection{What to Improve?}
\begin{enumerate}
    \item \textbf{Credit Assignment:} The preference provided by humans is on a trajectory level. Hence the learned reward model can only compare responses on an entire level. Is there a way to assign credit to different tokens or part of tokens? A known fact in RL is dense reward problems are much easier to learn, though they do not necessarily outperform the sparse reward settings.~\cite{plappert2018multi} (because of local minima, again, a reward engineering problem)
    \item \textbf{Algorithmic Design:} RL algorithms are seldom designed in a way that assumes knowing the dynamics model. But in LLM alignment, the actions are actually generated in an auto-regressive manner. Is there a more efficient and stable RL algorithm that works better than PPO in such a \textit{series generation} setting? This is a sort of Auto-Regressive MDP.
    \item \textbf{Prompting:} Is the prompting strategy optimized? Maybe the prompting strategy is not correct in getting the desired answer. Prompt optimization can definitely help improve the performance of LLMs. \textit{To address such a point, we introduce recent work on query-dependent prompt optimization~\cite{sun2023offline} in the next section, which also links RL and LLM.}
\end{enumerate}

\section{Prompting with Offline IRL:  Prompt Optimization is RL from AI Feedback}

\begin{figure}[h!]
    \centering
    \includegraphics[width=1.0\linewidth]{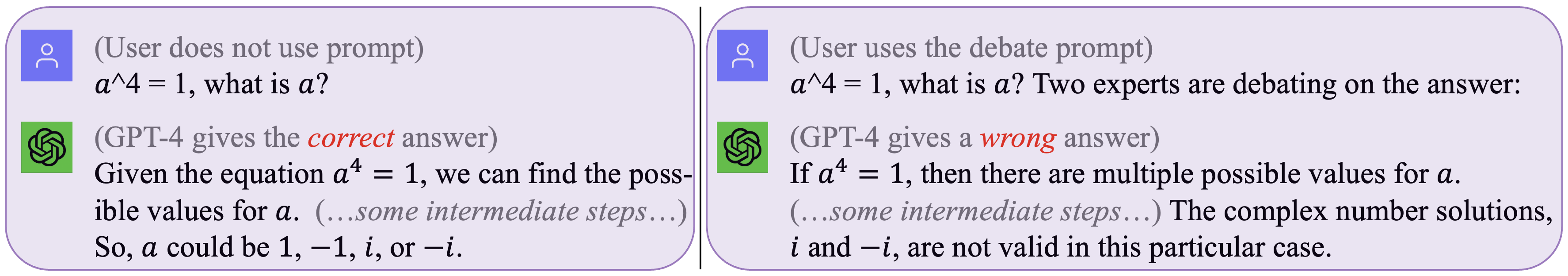}
    \caption{\small (From \citet{sun2023offline}.) A motivating example (\href{https://chat.openai.com/share/0f2d11b1-322a-4c47-a877-ad6fbace8179}{left}, \href{https://chat.openai.com/share/15870a47-93c7-4b98-96c8-af0516c0c999}{right}). No prompt is perfect that works for all queries. The optimal prompt is query-dependent. Yet the seeking of such prompts is hindered by 2 challenges. Prompt-OIRL~\cite{sun2023offline} optimizes prompt during inference on a query-dependent level effectively and cost-efficiently.}
    \label{fig:10_prompt_motiv}
\end{figure}

\subsection{The Query-Dependent Prompting Problem}
Out of the many attempts, \textit{prompting} --- a natural language prefix or instruction that explains how to complete the task --- stands out as a lightweight promising solution for eliciting the capabilities of LLMs without model parameter tuning. While the advances in zero-shot prompting strategies highlight the potential for finding effective query-independent solutions, its reliance on manual crafting efforts and the vast search space over natural language intensifies the difficulty in discovering effective prompts.
Moreover, as demonstrated in Figure~\ref{fig:10_prompt_motiv}, the optimal prompt is query dependent --- there is no perfect prompt that works for all queries.

\subsection{Prompt-OIRL: Prompt Evaluation and Optimization with Offline Inverse RL}

Prompt-OIRL is a novel approach grounded in offline inverse reinforcement learning, designed to reconcile effective and cost-efficient query-dependent prompt evaluation and optimization. This method leverages offline datasets from existing evaluations, utilizing Inverse-RL to craft a reward model tailored for offline, query-specific prompt evaluations. Prompt-OIRL offers several benefits: it forecasts prompt efficacy, minimizes costs, and explores the prompt space more effectively --- all at a query-dependent level. We validate our approach across various LLMs and arithmetic reasoning datasets, underscoring its viability as a formidable solution for query-dependent offline prompt evaluation and optimization.

\subsection{Potential Applications}
While Prompt-OIRL primarily centers on arithmetic reasoning tasks, we wish to underscore the versatility of Prompt-OIRL's insights for broader applications, especially where there exists a prompting demonstration dataset accompanied by ratings of the prompted responses. As a hypothetical approach to dataset construction with human annotators incorporated into the process, consider this: human annotators could employ LLMs to accomplish specific tasks. They might offer multiple prompts as instructions for the task, and the ensuing LLM responses can then be graded based on proficiency in executing the given task. In fact, these annotators could be everyday LLM users keen on evaluating diverse responses. We earmark this intriguing concept for subsequent exploration.